\documentclass[11pt,a4paper]{article}
\usepackage[hyperref]{acl2018}
\usepackage{times}
\usepackage{latexsym}
\usepackage{enumitem}

\usepackage{url}

\aclfinalcopy 
\def\aclpaperid{1249} 

\usepackage{yfonts}
\usepackage{graphicx} 
\usepackage{subcaption}

\usepackage[linesnumbered,ruled,vlined]{algorithm2e}

\SetCommentSty{mycommfont}

\usepackage{stackengine}  
\usepackage{tikz}

\usepackage{verbatim}
\usepackage{etoolbox}
\usepackage{framed}
\usepackage{xcolor}
\usepackage{color}
\usepackage{amsmath}

\usepackage{multirow}
\usepackage{arydshln}

\title{Stack-Pointer Networks for Dependency Parsing}

\author{
Xuezhe Ma \\ Carnegie Mellon University \\ {\tt xuezhem@cs.cmu.edu}
\And Zecong Hu\Thanks{Work done while at Carnegie Mellon University.} \\ Tsinghua University \\ {\tt huzecong@gmail.com}
\And Jingzhou Liu \\ Carnegie Mellon University \\ {\tt liujingzhou@cs.cmu.edu}
\AND 
Nanyun Peng \\ University of Southern California \\ {\tt npeng@isi.edu} 
\And Graham Neubig \and Eduard Hovy \\ Carnegie Mellon University \\ {\tt \{gneubig, ehovy\}@cs.cmu.edu}
}

\date{}

\usepackage{etoolbox}
\makeatletter
\patchcmd\@combinedblfloats{\box\@outputbox}{\unvbox\@outputbox}{}{%
  \errmessage{\noexpand\@combinedblfloats could not be patched}%
}%
\makeatother

\begin{document}

\maketitle

\begin{abstract}
We introduce a novel architecture for dependency parsing: \emph{stack-pointer networks} (\textbf{\textsc{StackPtr}}). 
Combining pointer networks~\citep{vinyals2015pointer} with an internal stack, the proposed model first reads and encodes the whole sentence, then builds the dependency tree top-down (from root-to-leaf) in a depth-first fashion. 
The stack tracks the status of the depth-first search and the pointer networks select one child for the word at the top of the stack at each step. 
The \textsc{StackPtr} parser benefits from the information of the whole sentence and all previously derived subtree structures, and removes the left-to-right restriction in classical transition-based parsers.
Yet, the number of steps for building any (including non-projective) parse tree is linear in the length of the sentence just as other transition-based parsers, yielding an efficient decoding algorithm with $O(n^2)$ time complexity. 
We evaluate our model on 29 treebanks spanning 20 languages and different dependency annotation schemas, and achieve state-of-the-art performance on 21 of them.
\end{abstract}

\section{Introduction}
\label{sec:intro}
Dependency parsing, which predicts the existence and type of linguistic dependency relations between words, is a first step towards deep language understanding. Its importance is widely recognized in the natural language processing (NLP) community, with it benefiting a wide range of NLP applications, such as coreference resolution~\citep{ng:2010:ACL,durrett-klein:2013:EMNLP,ma-hovy:2016:NAACL}, sentiment analysis~\citep{tai-socher-manning:2015:ACL-IJCNLP}, machine translation~\citep{bastings-EtAl:2017:EMNLP2017}, information extraction~\citep{nguyen:2009:EMNLP,angeli:2015:ACL,peng2017cross}, 
word sense disambiguation~\citep{fauceglia-EtAl:2015:EVENTS}, and low-resource languages processing~\citep{mcdonald-EtAl:2013:Short,ma-xia:2014:P14-1}.
There are two dominant approaches to dependency parsing~\citep{Buchholz:2006,nivre:CoNLL2007}: local and greedy \emph{transition-based} algorithms~\citep{yamada2003statistical,Nivre:2004,zhang-nivre:2011,chen-manning:EMNLP2014},  and the globally optimized \emph{graph-based} algorithms~\citep{eisner1996three,McDonald:2005,McDonald:2005b,Koo:2010}.

Transition-based dependency parsers read words sequentially (commonly from left-to-right) and build dependency trees incrementally by making series of multiple choice  decisions. 
The advantage of this formalism is that the number of operations required to build any projective parse tree is linear with respect to the length of the sentence. 
The challenge, however, is that the decision made at each step is based on local information, leading to error propagation and worse performance compared to graph-based parsers on root and long dependencies~\citep{mcdonald2011analyzing}.
Previous studies have explored solutions to address this challenge.
Stack LSTMs~\citep{dyer-EtAl:2015:ACL-IJCNLP,ballesteros-dyer-smith:2015:EMNLP,ballesteros-EtAl:2016:EMNLP2016} are capable of learning representations of the parser state that are sensitive to the complete contents of the parser's state. 
\citet{andor-EtAl:2016:P16-1} proposed a globally normalized transition model to replace the locally normalized classifier.
However, the parsing accuracy is still behind state-of-the-art graph-based parsers~\citep{dozat2017:ICLR}. 

Graph-based dependency parsers, on the other hand, learn scoring functions for parse trees and perform exhaustive search over all possible trees for a sentence to find the globally highest scoring tree. 
Incorporating this global search algorithm with distributed representations learned from neural networks, neural graph-based parsers~\citep{TACL885,wang-chang:2016:P16-1,kuncoro-EtAl:2016:EMNLP2016,dozat2017:ICLR} have achieved the state-of-the-art accuracies on a number of treebanks in different languages.
Nevertheless, these models, while accurate, are usually slow (e.g. decoding is $O(n^3)$ time complexity for first-order models~\cite{McDonald:2005,McDonald:2005b} and higher polynomials for higher-order models~\citep{mcdonald2006online,Koo:2010,ma2012probabilistic,ma-zhao:2012:POSTERS}).

In this paper, we propose a novel neural network architecture for dependency parsing, \emph{stack-pointer networks} (\textbf{\textsc{StackPtr}}).
\textsc{StackPtr} is a transition-based architecture, with the corresponding asymptotic efficiency, but still maintains a global view of the sentence that proves essential for achieving competitive accuracy.
Our \textsc{StackPtr} parser has a pointer network~\citep{vinyals2015pointer} as its backbone, and is equipped with an internal stack to maintain the order of head words in tree structures. 
The \textsc{StackPtr} parser performs parsing in an incremental, top-down, depth-first fashion; at each step, it generates an arc by assigning a child for the head word at the top of the internal stack. 
This architecture makes it possible to capture information from the whole sentence and all the previously derived subtrees, while maintaining a number of parsing steps linear in the sentence length.

We evaluate our parser on 29 treebanks across 20 languages and different dependency annotation schemas, and achieve state-of-the-art performance on 21 of them. 
The contributions of this work are summarized as follows: 
\begin{enumerate}[topsep=0pt,itemsep=-1.5ex,label=(\roman*)]
\item We propose a neural network architecture for dependency parsing that is simple, effective, and efficient. 
\item Empirical evaluations on benchmark datasets over 20 languages show that our method achieves state-of-the-art performance on 21 different treebanks\footnote{Source code is publicly available at \url{https://github.com/XuezheMax/NeuroNLP2}}.
\item Comprehensive error analysis is conducted to compare the proposed method to a strong graph-based baseline using biaffine attention~\citep{dozat2017:ICLR}.
\end{enumerate}

\section{Background}
\label{sec:model}
We first briefly describe the task of dependency parsing, setup the notation, and review Pointer Networks~\citep{vinyals2015pointer}.

\begin{figure*}
\centering
\stackinset{l}{}{b}{}{
	\begin{subfigure}{0.3\linewidth}
    	\tikz{\node[draw,dashed]{
        	\includegraphics[width=\linewidth]{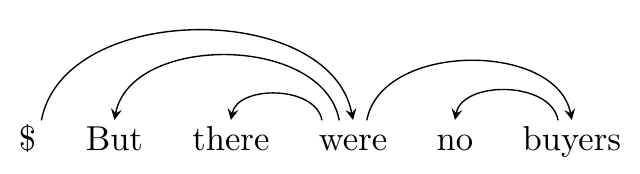}
        }}
		\renewcommand{\thesubfigure}{a} 
		\caption{}
	\end{subfigure}%
}{
	\begin{subfigure}{\linewidth}
		\includegraphics[width=\linewidth]{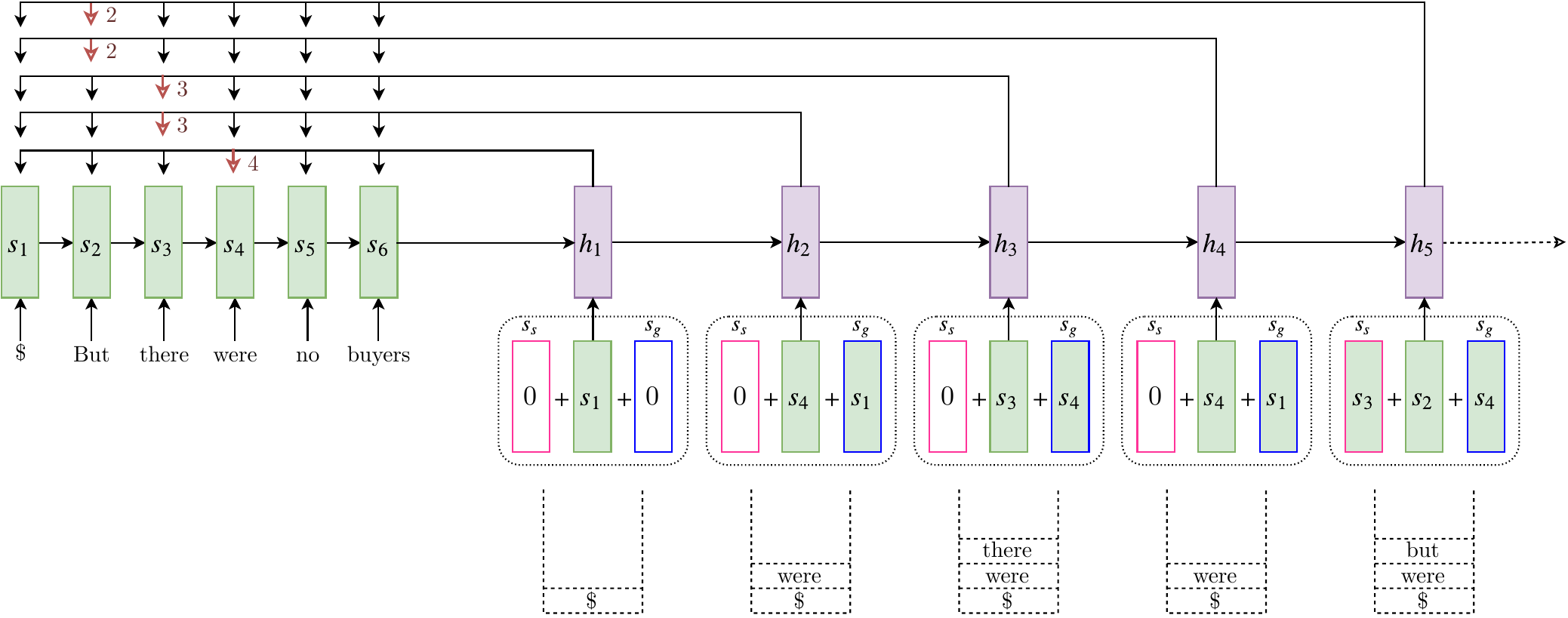}
		\renewcommand{\thesubfigure}{b}
		\caption{\label{fig:model}}
	\end{subfigure}
}
\caption{Neural architecture for the \textsc{StackPtr} network, together with the decoding procedure of an example sentence. 
The BiRNN of the encoder is elided for brevity. 
For the inputs of decoder at each time step, vectors in red and blue boxes indicate the sibling and grandparent.}\label{fig:architecture}
\end{figure*}

\subsection{Dependency Parsing and Notations}
Dependency trees represent syntactic relationships between words in the sentences through labeled directed edges between head words 
and their dependents. 
Figure~\ref{fig:architecture} (a) shows a dependency tree for the sentence, ``But there were no buyers''.

In this paper, we will use the following notation: 

\textbf{Input}: $\mathbf{x} = \{w_1, \ldots, w_n \}$ represents 
a generic sentence, where $w_i$ is the $i$th word.

\textbf{Output}: $\mathbf{y} = \{ p_1, p_2, \cdots, p_k \}$ represents a generic (possibly non-projective) dependency tree, where each path $p_i = \$, w_{i,1}, w_{i,2}, \cdots, w_{i,l_i}$ is a sequence of words from the root to a leaf.
``\$'' is an universal virtual root that is added to each tree.

\textbf{Stack}: $\sigma$ denotes a stack configuration, which is a sequence of words. We use $\sigma|w$ to represent a stack configuration that pushes word $w$ into the stack $\sigma$.

\textbf{Children}: $\mathrm{ch}(w_i)$ denotes the list of all the children (modifiers) of word $w_i$.

\subsection{Pointer Networks}\label{subsec:ptr}
Pointer Networks (\textsc{Ptr-Net}) \citep{vinyals2015pointer} are a variety of neural network capable of learning the conditional probability of an output sequence with elements that are discrete tokens corresponding to positions in an input sequence.
This model cannot be trivially expressed by standard sequence-to-sequence networks~\citep{sutskever2014sequence} due to the variable number of input positions in each sentence. 
\textsc{Ptr-Net} solves the problem by using attention~\citep{bahdanau2015,luong-pham-manning:2015:EMNLP} as a pointer to select a member of the input sequence as the output.

Formally, the words of the sentence $\mathbf{x}$ are fed one-by-one into the encoder (a multiple-layer bi-directional RNN), producing a sequence of \emph{encoder hidden states $s_i$}. 
At each time step $t$, the decoder (a uni-directional RNN) receives the input from last step and outputs \emph{decoder hidden state $h_t$}. 
The \emph{attention vector $a^t$} is calculated as follows: 
\begin{equation}\label{eq:attention}
\begin{array}{rcl}
e^{t}_{i} & = & \mathit{score}(h_t, s_i) \\
a^t & = & \mathit{softmax}(e^t)
\end{array}
\end{equation}
where $\mathit{score}(\cdot, \cdot)$ is the \emph{attention scoring function}, which has several variations such as dot-product, concatenation, and biaffine~\citep{luong-pham-manning:2015:EMNLP}. 
\textsc{Ptr-Net} regards the attention vector $a^t$ as a probability distribution over the source words, i.e. it uses $a^t_i$ as pointers to select the input elements.

\section{Stack-Pointer Networks}
\label{subsec:stackptr}

\subsection{Overview}
Similarly to \textsc{Ptr-Net}, \textsc{StackPtr} first reads the whole sentence and encodes each word into the encoder hidden state $s_i$. 
The internal stack $\sigma$ is always initialized with the root symbol \$. 
At each time step $t$, the decoder receives the input vector corresponding to the top element of the stack $\sigma$ (the head word $w_p$ where $p$ is the word index), generates the hidden state $h_t$, and computes the attention vector $a^t$ using Eq.~\eqref{eq:attention}. 
The parser chooses a specific position $c$ according to the attention scores in $a^t$ to generate a new dependency arc $(w_h, w_c)$ by selecting $w_c$ as a child of $w_h$. Then the parser pushes $w_c$ onto the stack, i.e. $\sigma \rightarrow \sigma|w_c$, and goes to the next step.
At one step if the parser points $w_h$ to itself, i.e. $c = h$, it indicates that all children of the head word $w_h$ have already been selected. Then the parser goes to the next step by popping $w_h$ out of $\sigma$.

At test time, in order to guarantee a valid dependency tree containing all the words in the input sentences exactly once, the decoder maintains a list of ``available'' words. At each decoding step, the parser selects a child for the current head word, and removes the child from the list of available words to make sure that it cannot be selected as a child of other head words.

For head words with multiple children, it is possible that there is more than one valid selection for each time step. 
In order to define a deterministic decoding process to make sure that there is only one ground-truth choice at each step (which is necessary for simple maximum likelihood estimation), a predefined order for each $\mathrm{ch}(w_i)$ needs to be introduced. 
The predefined order of children can have different alternatives, such as left-to-right or inside-out\footnote{Order the children by the distances to the head word on the left side, then the right side.}. 
In this paper, we adopt the inside-out order\footnote{We also tried left-to-right order which obtained worse parsing accuracy than inside-out.} since it enables us to utilize second-order \emph{sibling} information, which has been proven beneficial for parsing performance~\citep{mcdonald2006online,Koo:2010} (see \S~\ref{subsec:higher-order} for details). Figure~\ref{fig:architecture} (b) depicts the architecture of \textsc{StackPtr} and the decoding procedure for the example sentence in Figure~\ref{fig:architecture}~(a).

\subsection{Encoder}\label{subsec:encoder}
The encoder of our parsing model is based on the bi-directional LSTM-CNN architecture (BLSTM-CNNs)~\citep{TACL792,ma-hovy:2016:P16-1} where CNNs encode character-level information of a word into its character-level representation and BLSTM models context information of each word. 
Formally, for each word, the CNN, with character embeddings as inputs, encodes the character-level representation. 
Then the character-level representation vector is concatenated with the word embedding vector to feed into the BLSTM network.
To enrich word-level information, we also use POS embeddings.
Finally, the encoder outputs a sequence of hidden states $s_i$.

\subsection{Decoder}\label{subsec:decoder}
The decoder for our parser is a uni-directional LSTM. Different from previous work \citep{bahdanau2015,vinyals2015pointer} which uses word embeddings of the previous word as the input to the decoder, our decoder receives the encoder hidden state vector ($s_i$) of the top element in the stack $\sigma$ (see Figure~\ref{fig:architecture} (b)).
Compared to word embeddings, the encoder hidden states contain more contextual information, benefiting both the training and decoding procedures. 
The decoder produces a sequence of decoder hidden states $h_i$, one for each decoding step.

\subsection{Higher-order Information}\label{subsec:higher-order}
As mentioned before, our parser is capable of utilizing higher-order information.
In this paper, we incorporate two kinds of higher-order structures --- \emph{grandparent} and \emph{sibling}.
A sibling structure is a head word with two successive modifiers, and a grandparent structure is a pair of dependencies connected head-to-tail:
\begin{figure}[h]
\centering
\includegraphics[scale=0.9]{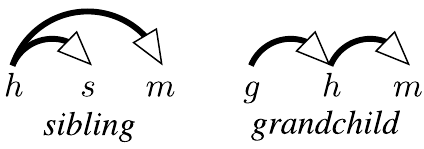}
\end{figure}

To utilize higher-order information, the decoder's input at each step is the sum of the encoder hidden states of three words:
\begin{displaymath}
\beta_t = s_h + s_g + s_s
\end{displaymath}
where $\beta_t$ is the input vector of decoder at time $t$ and $h, g, s$ are the indices of the head word and its grandparent and sibling, respectively. 
Figure~\ref{fig:architecture} (b) illustrates the details.
Here we use the element-wise sum operation instead of concatenation because it does not increase the dimension of the input vector $\beta_t$, thus introducing no additional model parameters.

\subsection{Biaffine Attention Mechanism}\label{subsec:attention}
For attention score function (Eq.~\eqref{eq:attention}), we adopt the biaffine attention mechanism~\citep{luong-pham-manning:2015:EMNLP,dozat2017:ICLR}:
\begin{displaymath}
e^{t}_{i} = h_t^T \mathbf{W} s_i + \mathbf{U}^T h_t + \mathbf{V}^T s_i + \mathbf{b}
\end{displaymath}
where $\mathbf{W}, \mathbf{U}, \mathbf{V}$, $\mathbf{b}$ are parameters, denoting the weight matrix of the bi-linear term, the two weight vectors of the linear terms, and the bias vector. 

As discussed in \newcite{dozat2017:ICLR}, applying a multilayer perceptron (MLP) to the output vectors of the BLSTM before the score function can both reduce the dimensionality and overfitting of the model. We follow this work by using a one-layer perceptron to $s_i$ and $h_i$ with elu~\cite{clevert2015fast} as its activation function.

Similarly, the dependency label classifier also uses a biaffine function to score each label, given the head word vector $h_t$ and child vector $s_i$ as inputs.
Again, we use MLPs to transform $h_t$ and $s_i$ before feeding them into the classifier.

\subsection{Training Objectives} 
\label{subsec:prob}
The \textsc{StackPtr} parser is trained to optimize the probability of the dependency trees given sentences: $P_{\theta}(\mathbf{y}|\mathbf{x})$, which can be factorized as:
{\setlength\arraycolsep{2pt}
\begin{equation}\label{eq:prob}
\begin{array}{rcl}
P_{\theta}(\mathbf{y}|\mathbf{x}) & = & \prod\limits_{i=1}^{k} P_{\theta}(p_i|p_{<i},\mathbf{x}) \\
 & = & \prod\limits_{i=1}^{k} \prod\limits_{j=1}^{l_i} P_{\theta}(c_{i,j}|c_{i,<j}, p_{<i},\mathbf{x}),
\end{array}
\end{equation}}%
where $\theta$ represents model parameters. $p_{<i}$ denotes the preceding paths that have already been generated. 
$c_{i,j}$ represents the $j$th word in $p_i$ and $c_{i,<j}$ denotes all the proceeding words on the path $p_i$. 
Thus, the \textsc{StackPtr} parser is an autoregressive model, like sequence-to-sequence models, but it factors the distribution according to a top-down tree structure as opposed to a left-to-right chain.
We define $P_{\theta}(c_{i,j}|c_{i,<j}, p_{<i},\mathbf{x}) = a^t$,
where attention vector $a^t$ (of dimension $n$) is used as the distribution over the indices of words in a sentence.

\paragraph{Arc Prediction}
Our parser is trained by optimizing the conditional likelihood in Eq~\eqref{eq:prob}, which is implemented as the cross-entropy loss.

\paragraph{Label Prediction}
We train a separated multi-class classifier in parallel to predict the dependency labels. Following \citet{dozat2017:ICLR}, the classifier takes the information of the head word and its child as features. The label classifier is trained simultaneously with the parser by optimizing the sum of their objectives. 

\subsection{Discussion}\label{subsec:discuss}
\paragraph{Time Complexity.} The number of decoding steps to build a parse tree for a sentence of length $n$ is $2n-1$, linear in $n$. 
Together with the attention mechanism (at each step, we need to compute the attention vector $a^t$, whose runtime is $O(n)$), the time complexity of decoding algorithm is $O(n^2)$, which is more efficient than graph-based parsers that have $O(n^3)$ or worse complexity when using dynamic programming or maximum spanning tree (MST) decoding algorithms.
\paragraph{Top-down Parsing.} When humans comprehend a natural language sentence, they arguably do it in an incremental, left-to-right manner.
However, when humans consciously annotate a sentence with syntactic structure, they rarely ever process in fixed left-to-right order.
Rather, they start by reading the whole sentence, then seeking the main predicates, jumping back-and-forth over the sentence and recursively proceeding to the sub-tree structures governed by certain head words. 
Our parser follows a similar kind of annotation process: starting from reading the whole sentence, and processing in a top-down manner by finding the main predicates first and only then search for sub-trees governed by them. When making latter decisions, the parser has access to the entire structure built in earlier steps.

\subsection{Implementation Details}\label{subsec:opt}
\paragraph{Pre-trained Word Embeddings.} For all the parsing models in different languages, we initialize word vectors with pretrained word embeddings. 
For Chinese, Dutch, English, German and Spanish, we use the structured-skipgram~\cite{ling-EtAl:2015:NAACL-HLT} embeddings. 
For other languages we use Polyglot embeddings~\cite{polyglot:2013:CoNLL}. 

\paragraph{Optimization.} Parameter optimization is performed with the Adam optimizer~\cite{kingma2014adam} with $\beta_1 = \beta_2 = 0.9$.
We choose an initial learning rate of $\eta_0 = 0.001$. The learning rate $\eta$ is annealed by multiplying a fixed decay rate $\rho = 0.75$ when parsing performance stops increasing on validation sets.
To reduce the effects of ``gradient exploding'', we use gradient clipping of $5.0$~\cite{pascanu2013}.

\paragraph{Dropout Training.} To mitigate overfitting, we apply dropout~\cite{srivastava2014dropout,iclr2017:ma:dropout}.
For BLSTM, we use recurrent dropout~\cite{gal2016dropout:rnn} with a drop rate of 0.33 between hidden states and 0.33 between layers.
Following \citet{dozat2017:ICLR}, we also use embedding dropout with a rate of 0.33 on all word, character, and POS embeddings.

\paragraph{Hyper-Parameters.}  
Some parameters are chosen from those reported in \citet{dozat2017:ICLR}.
We use the same hyper-parameters across the models on different treebanks and languages, due to time constraints.
The details of the chosen hyper-parameters for all experiments are summarized in Appendix A.

\begin{figure*}[t]
\centering
\includegraphics[width=\linewidth]{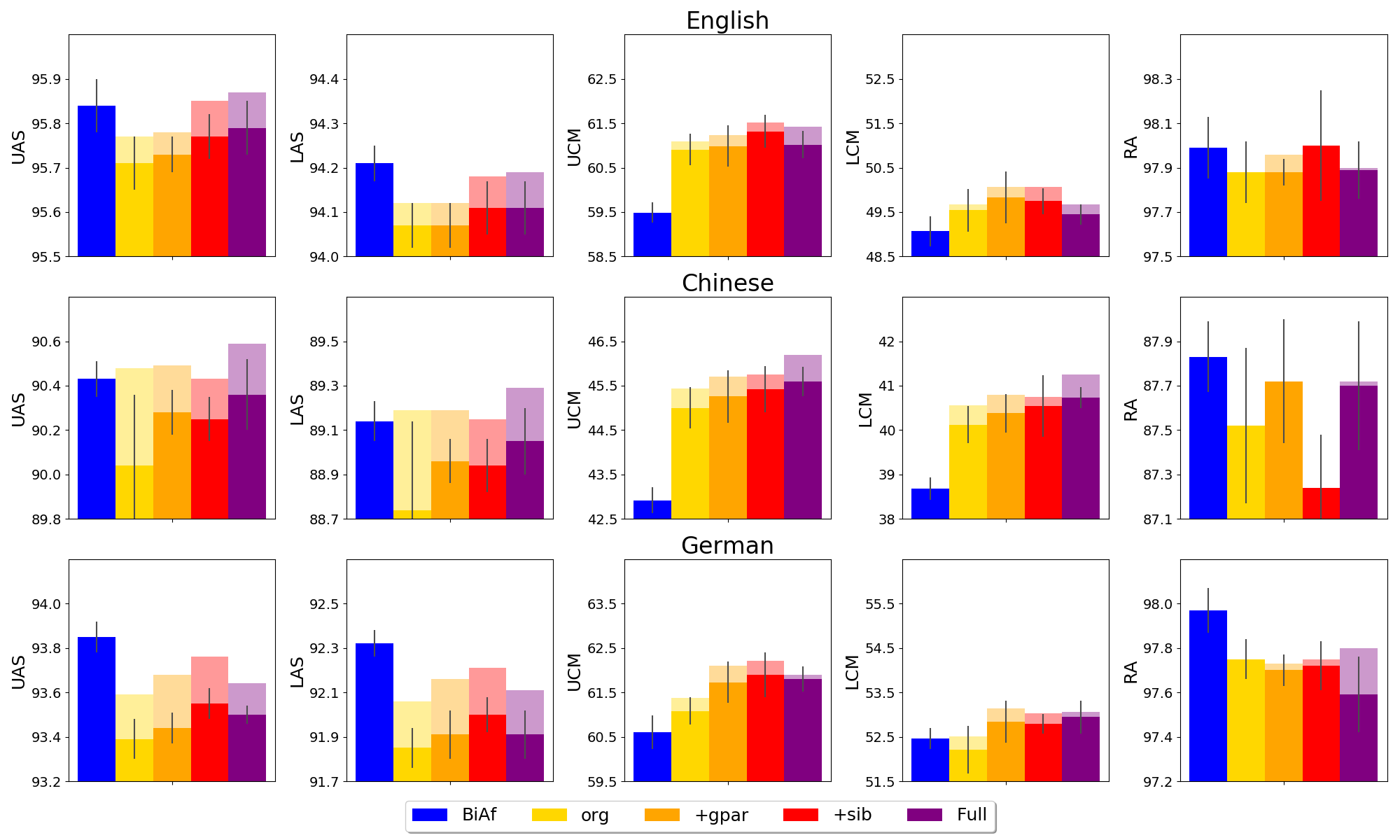}
\caption{Parsing performance of different variations of our model on the test sets for three languages, together with baseline \textsc{BiAF}. 
For each of our \textsc{StackPtr} models, we perform decoding with beam size equal to $1$ and $10$. 
The improvements of decoding with beam size $10$ over $1$ are presented by stacked bars with light colors.}
\label{fig:main:results}
\vspace{-.5cm}
\end{figure*}
			
\section{Experiments}\label{sec:experiment}
\subsection{Setup}
We evaluate our \textsc{StackPtr} parser mainly on three treebanks: the English Penn Treebank (PTB version 3.0)~\citep{Marcus:1993}, the Penn Chinese Treebank (CTB version 5.1)~\cite{xue2002building}, and the German CoNLL 2009 corpus~\cite{hajivc2009conll}.
We use the same experimental settings as \citet{kuncoro-EtAl:2016:EMNLP2016}.

To make a thorough empirical comparison with previous studies, we also evaluate our system on treebanks  from CoNLL shared task and the Universal Dependency (UD) Treebanks\footnote{\url{http://universaldependencies.org/}}. 
For the CoNLL Treebanks, we use the English treebank from CoNLL-2008 shared task~\cite{surdeanu2008conll} and all 13 treebanks from CoNLL-2006 shared task~\cite{Buchholz:2006}.
The experimental settings are the same as \citet{ma-hovy:2015:EMNLP}.
For UD Treebanks, we select 12 languages. The details of the treebanks and experimental settings are in \S~\ref{subsec:experiment:other} and Appendix B.

\paragraph{Evaluation Metrics} Parsing performance is measured with five metrics: unlabeled attachment score (UAS), labeled attachment score (LAS), unlabeled complete match (UCM), labeled complete match (LCM), and root accuracy (RA).
Following previous work~\citep{kuncoro-EtAl:2016:EMNLP2016,dozat2017:ICLR}, we report results excluding punctuations for Chinese and English.
For each experiment, we report the mean values with corresponding standard deviations over 5 repetitions.

\paragraph{Baseline} For fair comparison of the parsing performance, we re-implemented the graph-based Deep Biaffine (\textsc{BiAF}) parser~\citep{dozat2017:ICLR}, which achieved state-of-the-art results on a wide range of languages. 
Our re-implementation adds character-level information using the same LSTM-CNN encoder  as our model (\S~\ref{subsec:encoder}) to the original \textsc{BiAF} model, which boosts its performance on all languages.

\subsection{Main Results}
We first conduct experiments to demonstrate the effectiveness of our neural architecture by comparing with the strong baseline \textsc{BiAF}. 
We compare the performance of four variations of our model with different decoder inputs --- \textsf{Org}, \textsf{+gpar}, \textsf{+sib} and \textsf{Full} --- where 
the \textsf{Org} model utilizes only the encoder hidden states of head words, while the \textsf{+gpar} and \textsf{+sib} models augments the original one with grandparent and sibling information, respectively. 
The \textsf{Full} model includes all the three information as inputs.

Figure~\ref{fig:main:results} illustrates the performance (five metrics) of different variations of our \textsc{StackPtr} parser together with the results of baseline \textsc{BiAF} re-implemented by us, on the test sets of the three languages. 
On UAS and LAS, the \textsf{Full} variation of \textsc{StackPtr} with decoding beam size 10 outperforms \textsc{BiAF} on Chinese, and obtains competitive performance on English and German. An interesting observation is that the \textsf{Full} model achieves the best accuracy on English and Chinese, while performs slightly worse than \textsf{+sib} on German. 
This shows that the importance of higher-order information varies in languages.
On LCM and UCM, \textsc{StackPtr} significantly outperforms \textsc{BiAF} on all languages, showing the superiority of our parser on complete sentence parsing. 
The results of our parser on RA are slightly worse than \textsc{BiAF}.
More details of results are provided in Appendix C.

\begin{table*}[t]
\centering
{\small
\begin{tabular}[t]{l|c|ll|ll|ll}
\hline
 & & \multicolumn{2}{c|}{\textbf{English}} & \multicolumn{2}{c|}{\textbf{Chinese}} & \multicolumn{2}{c}{\textbf{German}} \\
 \cline{3-8}
\textbf{System} & & UAS & LAS & UAS & LAS & UAS & LAS \\
\hline
\citet{chen-manning:EMNLP2014} & T & 91.8 & 89.6 & 83.9 & 82.4 & -- & -- \\
\citet{ballesteros-dyer-smith:2015:EMNLP} & T & 91.63 & 89.44 & 85.30 & 83.72 & 88.83 & 86.10 \\
\citet{dyer-EtAl:2015:ACL-IJCNLP} & T & 93.1 & 90.9 & 87.2 & 85.7 & -- & -- \\
\citet{bohnet-nivre:2012:EMNLP-CoNLL} & T & 93.33 & 91.22 & 87.3 & 85.9 & 91.4 & 89.4 \\
\citet{ballesteros-EtAl:2016:EMNLP2016} & T & 93.56 & 91.42 & 87.65 & 86.21 & -- & -- \\
\citet{TACL885} & T & 93.9 & 91.9 & 87.6 & 86.1 & -- & -- \\
\citet{weiss:2015:ACL} & T & 94.26 & 92.41 & -- & -- & -- & -- \\
\citet{andor-EtAl:2016:P16-1} & T & 94.61 & 92.79 & -- & -- & 90.91 & 89.15 \\
\hline
\citet{TACL885} & G & 93.1 & 91.0 & 86.6 & 85.1 & -- & -- \\
\citet{wang-chang:2016:P16-1} & G & 94.08 & 91.82 & 87.55 & 86.23 & -- & -- \\
\citet{cheng-EtAl:2016:EMNLP2016} & G & 94.10 & 91.49 & 88.1 & 85.7 & -- & -- \\
\citet{kuncoro-EtAl:2016:EMNLP2016} & G & 94.26 & 92.06 & 88.87 & 87.30 & 91.60 & 89.24 \\
\citet{ma-hovy:2017:I17-1} & G & 94.88 & 92.98 & 89.05 & 87.74 & 92.58 & 90.54 \\
\textsc{BiAF}: \citet{dozat2017:ICLR} & G & 95.74 & 94.08 & 89.30 & 88.23 & 93.46 & 91.44 \\
\textsc{BiAF}: re-impl & G & 95.84 & \textbf{94.21} & 90.43 & 89.14 & \textbf{93.85} & \textbf{92.32} \\
\hline
\textsc{StackPtr}: \textsf{Org} & T & 95.77 & 94.12 & 90.48 & 89.19 & 93.59 & 92.06 \\
\textsc{StackPtr}: \textsf{+gpar} & T & 95.78 & 94.12 & 90.49 & 89.19 & 93.65 & 92.12 \\
\textsc{StackPtr}: \textsf{+sib}  & T & 95.85 & 94.18 & 90.43 & 89.15 & 93.76 & 92.21 \\
\textsc{StackPtr}: \textsf{Full}  & T & \textbf{95.87} & 94.19 & \textbf{90.59} & \textbf{89.29} & 93.65 & 92.11 \\
\hline
\end{tabular}}
\caption{UAS and LAS of four versions of our model on test sets for three languages, together with top-performing parsing systems. 
``T'' and ``G'' indicate transition- and graph-based models, respectively. 
For \textsc{BiAF}, we provide the original results reported in \citet{dozat2017:ICLR} and our re-implementation.
For \textsc{StackPtr} and our re-implementation of BiAF, we report the average over 5 runs.}
\label{tab:main:comparison}
\end{table*}

\begin{figure*}[t]
\centering
\begin{subfigure}{0.19\linewidth}
\includegraphics[width=\linewidth]{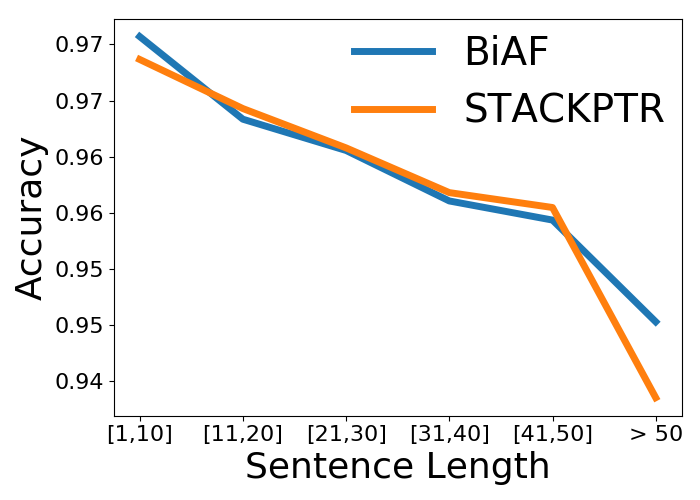}
\caption{\label{fig:sent_length}}
\end{subfigure}%
\hfill
\begin{subfigure}{0.39\linewidth}
\includegraphics[width=\linewidth]{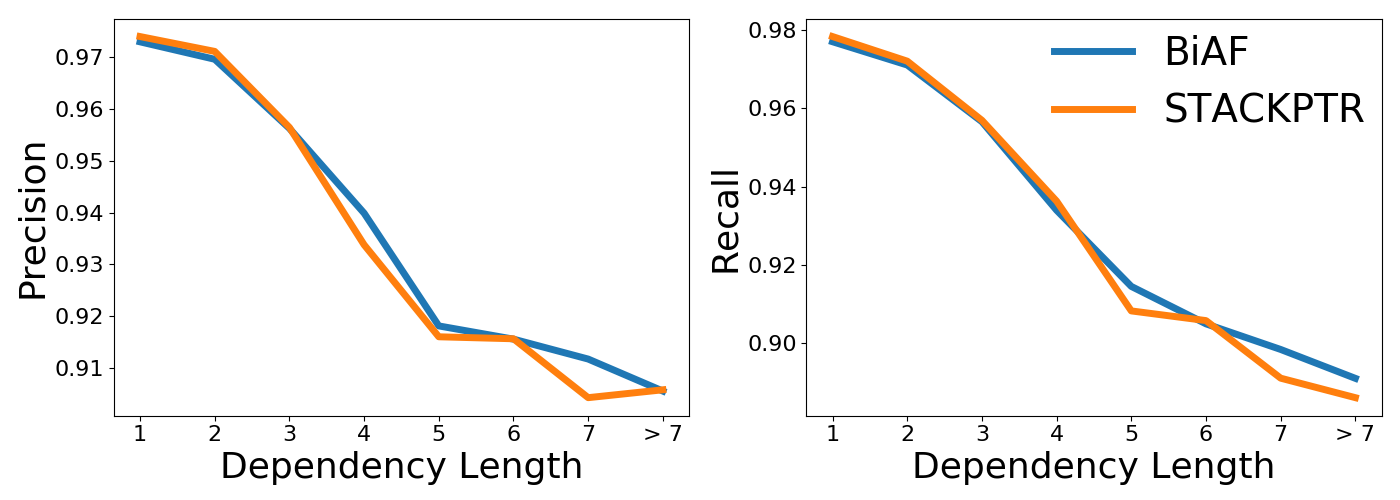}
\caption{\label{fig:dep_length}}
\end{subfigure}
\hfill
\begin{subfigure}{0.39\linewidth}
\includegraphics[width=\linewidth]{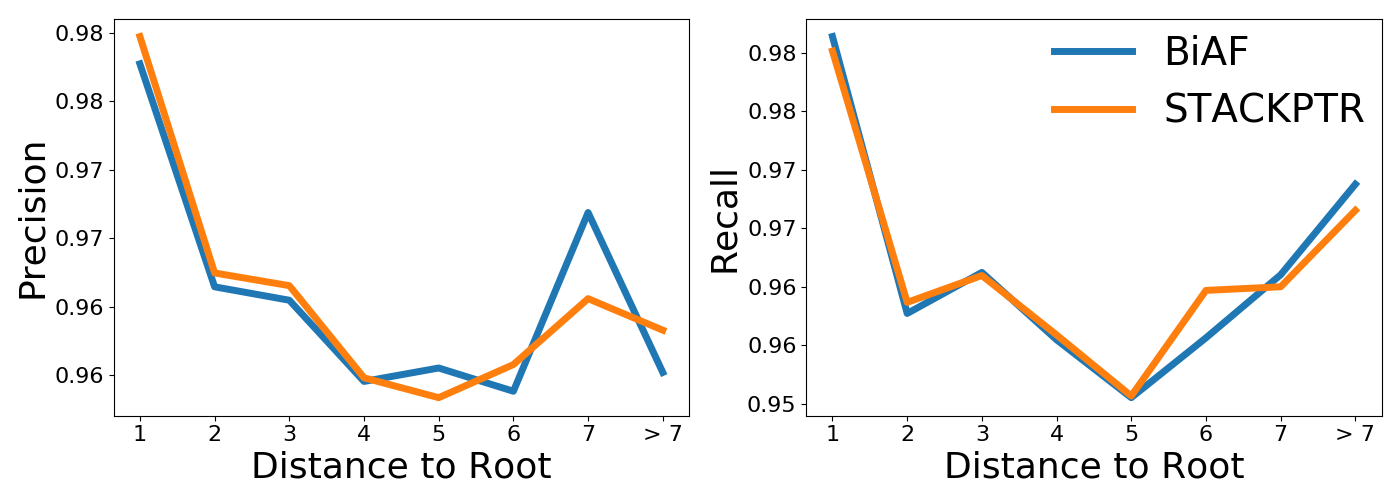}
\caption{\label{fig:root_dist}}
\end{subfigure}
\caption{Parsing performance of \textsc{BiAF} and \textsc{StackPtr} parsers relative to length and graph factors.}\label{fig:error_analysis}
\end{figure*}

\begin{table}[!t]
\centering
\setlength{\tabcolsep}{2pt}
{\small
\begin{tabular}[t]{l|c:c:c:c}
\hline
\textbf{POS} & UAS & LAS & UCM & LCM \\
\hline
\textsf{Gold} & 96.12$\pm$0.03 & 95.06$\pm$0.05 & 62.22$\pm$0.33 & 55.74$\pm$0.44 \\
\textsf{Pred} & 95.87$\pm$0.04 & 94.19$\pm$0.04 & 61.43$\pm$0.49 & 49.68$\pm$0.47 \\
\textsf{None} & 95.90$\pm$0.05 & 94.21$\pm$0.04 & 61.58$\pm$0.39 & 49.87$\pm$0.46 \\
\hline
\end{tabular}
}
\caption{Parsing performance on the test data of PTB with different versions of POS tags.}\label{tab:pos}
\end{table}

\subsection{Comparison with Previous Work}
Table~\ref{tab:main:comparison} illustrates the UAS and LAS of the four versions of our model (with decoding beam size 10) on the three treebanks, together with previous top-performing systems for comparison. 
Note that the results of \textsc{StackPtr} and our re-implementation of \textsc{BiAF} are the average of 5 repetitions instead of a single run.
Our \textsf{Full} model significantly outperforms all the transition-based parsers on all three languages, and achieves better results than most graph-based parsers.
Our re-implementation of \textsc{BiAF} obtains better performance than the original one in \citet{dozat2017:ICLR}, demonstrating the effectiveness of the character-level information.
Our model  achieves state-of-the-art performance on both UAS and LAS on Chinese, and best UAS on English. 
On German, the performance is competitive with \textsc{BiAF}, and significantly better than other models.

\begin{table*}[t]
\centering
{\small
\begin{tabular}[t]{l|c|c|c|c||cc}
\hline
 & \textsf{Bi-Att} & \textsf{NeuroMST} & \textsc{BiAF} & \textsc{StackPtr} & \multicolumn{2}{c}{\textbf{Best Published}} \\
 \cline{2-7}
 & UAS [LAS] & UAS [LAS] & UAS [LAS] & UAS [LAS] & UAS & LAS \\
\hline
ar  & 80.34 [68.58] & 80.80 [69.40] & 82.15$\pm$0.34 [71.32$\pm$0.36] & \textbf{83.04$\pm$0.29 [72.94$\pm$0.31]} & 81.12 & -- \\
bg  & 93.96 [89.55] & 94.28 [90.60] & 94.62$\pm$0.14 [\textbf{91.56$\pm$0.24}] & \textbf{94.66$\pm$0.10} [91.40$\pm$0.08] & 94.02 & -- \\
zh  & --            & 93.40 [90.10] & \textbf{94.05$\pm$0.27 [90.89$\pm$0.22]} & 93.88$\pm$0.24 [90.81$\pm$0.55] & 93.04 & -- \\
cs  & 91.16 [85.14] & 91.18 [85.92] & 92.24$\pm$0.22 [87.85$\pm$0.21] & \textbf{92.83$\pm$0.13 [88.75$\pm$0.16]} & 91.16 & 85.14 \\
da  & 91.56 [85.53] & 91.86 [87.07] & \textbf{92.80$\pm$0.26 [88.36$\pm$0.18]} & 92.08$\pm$0.15 [87.29$\pm$0.21] & 92.00 & -- \\
nl  & 87.15 [82.41] & 87.85 [84.82] & 90.07$\pm$0.18 [\textbf{87.24$\pm$0.17}] & \textbf{90.10$\pm$0.27} [87.05$\pm$0.26] & 87.39 & -- \\
en  & --            & 94.66 [92.52] & 95.19$\pm$0.05 [93.14$\pm$0.05] & \textbf{93.25$\pm$0.05 [93.17$\pm$0.05]} & 93.25 & -- \\
de  & 92.71 [89.80] & 93.62 [91.90] & 94.52$\pm$0.11 [93.06$\pm$0.11] & \textbf{94.77$\pm$0.05 [93.21$\pm$0.10]} & 92.71 & 89.80 \\
ja  & 93.44 [90.67] & \textbf{94.02 [92.60]} & 93.95$\pm$0.06 [92.46$\pm$0.07] & 93.38$\pm$0.08 [91.92$\pm$0.16] & 93.80 & -- \\
pt  & 92.77 [88.44] & 92.71 [88.92] & 93.41$\pm$0.08 [89.96$\pm$0.24] & \textbf{93.57$\pm$0.12 [90.07$\pm$0.20]} & 93.03 & -- \\
sl  & 86.01 [75.90] & 86.73 [77.56] & 87.55$\pm$0.17 [78.52$\pm$0.35] & \textbf{87.59$\pm$0.36 [78.85$\pm$0.53]} & 87.06 & -- \\
es  & 88.74 [84.03] & 89.20 [85.77] & 90.43$\pm$0.13 [87.08$\pm$0.14] & \textbf{90.87$\pm$0.26 [87.80$\pm$0.31]} & 88.75 & 84.03 \\
sv  & 90.50 [84.05] & 91.22 [86.92] & 92.22$\pm$0.15 [88.44$\pm$0.17] & \textbf{92.49$\pm$0.21 [89.01$\pm$0.22]} & 91.85 & 85.26 \\
tr  & 78.43 [66.16] & 77.71 [65.81] & \textbf{79.84$\pm$0.23 [68.63$\pm$0.29]} & 79.56$\pm$0.22 [68.03$\pm$0.15] & 78.43 & 66.16 \\
\hline
\end{tabular}
}
\caption{UAS and LAS on 14 treebanks from CoNLL shared tasks, together with several state-of-the-art parsers.
\textsf{Bi-Att} is the bi-directional attention based parser~\citep{cheng-EtAl:2016:EMNLP2016}, and \textsf{NeuroMST} is the neural MST parser~\citep{ma-hovy:2017:I17-1}.
``Best Published'' includes the most accurate parsers in term of UAS among \citet{koo-EtAl:2010:EMNLP}, \citet{martins-EtAl:2011:EMNLP1}, \citet{martins:2013:Short}, \citet{lei-EtAl:2014}, \citet{zhang-EtAl:2014:EMNLP20143}, \citet{zhang-mcdonald:2014}, \citet{pitler-mcdonald:2015}, and \citet{cheng-EtAl:2016:EMNLP2016}.}
\label{tab:conll}
\end{table*}

\subsection{Error Analysis}\label{subsec:analyze}
In this section, we characterize the errors made by \textsc{BiAF} and \textsc{StackPtr} by presenting a number of experiments that relate parsing errors to a set of linguistic and structural properties. 
For simplicity, we follow \citet{mcdonald2011analyzing} and report labeled parsing metrics (either accuracy, precision, or recall) for all experiments.

\subsubsection{Length and Graph Factors}
Following \citet{mcdonald2011analyzing}, we analyze parsing errors related to structural factors.
\paragraph{Sentence Length.}
Figure~\ref{fig:error_analysis} (a) shows the accuracy of both parsing models relative to sentence lengths.
Consistent with the analysis in \citet{mcdonald2011analyzing}, \textsc{StackPtr} tends to perform better on shorter sentences, which make fewer parsing decisions, significantly reducing the chance of error propagation.
\paragraph{Dependency Length.}
Figure~\ref{fig:error_analysis} (b) measures the precision and recall relative to dependency lengths.
While the graph-based \textsc{BiAF} parser still performs better for longer dependency arcs and transition-based \textsc{StackPtr} parser does better for shorter ones, the gap between the two systems is marginal, much smaller than that shown in \citet{mcdonald2011analyzing}. 
One possible reason is that, unlike traditional transition-based parsers that scan the sentence from left to right, \textsc{StackPtr} processes in a top-down manner, thus sometimes unnecessarily creating shorter dependency arcs first.

\paragraph{Root Distance.}
Figure~\ref{fig:error_analysis} (c) plots the precision and recall of each system for arcs of varying distance to the root. 
Different from the observation in \citet{mcdonald2011analyzing}, \textsc{StackPtr} does not show an obvious advantage on the precision for arcs further away from the root. 
Furthermore, the \textsc{StackPtr} parser does not have the tendency to over-predict root modifiers reported in \citet{mcdonald2011analyzing}. 
This behavior can be explained using the same reasoning as above: the fact that arcs further away from the root are usually constructed early in the parsing algorithm of traditional transition-based parsers is not true for the \textsc{StackPtr} parser.

\subsubsection{Effect of POS Embedding}
The only prerequisite information that our parsing model relies on is POS tags. 
With the goal of achieving an end-to-end parser, we explore the effect of POS tags on parsing performance.
We run experiments on PTB using our \textsc{StackPtr} parser with gold-standard and predicted POS tags, and without tags, respectively.
\textsc{StackPtr} in these experiments is the \textsf{Full} model with beam$=$10.  

Table~\ref{tab:pos}~gives results of the parsers with different versions of POS tags on the test data of PTB.
The parser with gold-standard POS tags significantly outperforms the other two parsers, showing that dependency parsers can still benefit from accurate POS information. 
The parser with predicted (imperfect) POS tags, however, performs even slightly worse than the parser without using POS tags.
It illustrates that an end-to-end parser that doesn't rely on POS information can obtain competitive (or even better) performance than parsers using imperfect predicted POS tags, even if the POS tagger is relative high accuracy (accuracy $> 97\%$ in this experiment on PTB).

\begin{table*}[t]
\centering
{\small\setlength{\tabcolsep}{5pt}
\begin{tabular}[t]{l|c:c|c:c||c:c|c:c}
\hline
 & \multicolumn{4}{c||}{\textbf{Dev}} & \multicolumn{4}{c}{\textbf{Test}} \\
 \cline{2-9}
 & \multicolumn{2}{c|}{\textsc{BiAF}} & \multicolumn{2}{c||}{\textsc{StackPtr}} & \multicolumn{2}{c|}{\textsc{BiAF}} & \multicolumn{2}{c}{\textsc{StackPtr}} \\
 \cline{2-9}
 & UAS & LAS & UAS & LAS & UAS & LAS & UAS & LAS \\
\hline
bg & 93.92$\pm$0.13 & 89.05$\pm$0.11 & \textbf{94.09$\pm$0.16} & \textbf{89.17$\pm$0.14} & 94.30$\pm$0.16 & \textbf{90.04$\pm$0.16} & \textbf{94.31$\pm$0.06} & 89.96$\pm$0.07 \\
ca & 94.21$\pm$0.05 & 91.97$\pm$0.06 & \textbf{94.47$\pm$0.02} & \textbf{92.51$\pm$0.05} & 94.36$\pm$0.06 & 92.05$\pm$0.07 & \textbf{94.47$\pm$0.02} & \textbf{92.39$\pm$0.02} \\
cs & 94.14$\pm$0.03 & 90.89$\pm$0.04 & \textbf{94.33$\pm$0.04} & \textbf{91.24$\pm$0.05} & 94.06$\pm$0.04 & 90.60$\pm$0.05 & \textbf{94.21$\pm$0.06} & \textbf{90.94$\pm$0.07} \\
de & 91.89$\pm$0.11 & 88.39$\pm$0.17 & \textbf{92.26$\pm$0.11} & \textbf{88.79$\pm$0.15} & 90.26$\pm$0.19 & 86.11$\pm$0.25 & \textbf{90.26$\pm$0.07} & \textbf{86.16$\pm$0.01} \\
en & \textbf{92.51$\pm$0.08} & \textbf{90.50$\pm$0.07} & 92.47$\pm$0.03 & 90.46$\pm$0.02 & 91.91$\pm$0.17 & 89.82$\pm$0.16 & \textbf{91.93$\pm$0.07} & \textbf{89.83$\pm$0.06} \\
es & 93.46$\pm$0.05 & 91.13$\pm$0.07 & \textbf{93.54$\pm$0.06} & \textbf{91.34$\pm$0.05} & 93.72$\pm$0.07 & 91.33$\pm$0.08 & \textbf{93.77$\pm$0.07} & \textbf{91.52$\pm$0.07} \\
fr & \textbf{95.05$\pm$0.04} & \textbf{92.76$\pm$0.07} & 94.97$\pm$0.04 & 92.57$\pm$0.06 & 92.62$\pm$0.15 & 89.51$\pm$0.14 & \textbf{92.90$\pm$0.20} & \textbf{89.88$\pm$0.23} \\
it & 94.89$\pm$0.12 & 92.58$\pm$0.12 & \textbf{94.93$\pm$0.09} & \textbf{92.90$\pm$0.10} & \textbf{94.75$\pm$0.12} & \textbf{92.72$\pm$0.12} & 94.70$\pm$0.07 & 92.55$\pm$0.09 \\
nl & 93.39$\pm$0.08 & 90.90$\pm$0.07 & \textbf{93.94$\pm$0.11} & \textbf{91.67$\pm$0.08} & 93.44$\pm$0.09 & 91.04$\pm$0.06 & \textbf{93.98$\pm$0.05} & \textbf{91.73$\pm$0.07} \\
no & 95.44$\pm$0.05 & 93.73$\pm$0.05 & \textbf{95.52$\pm$0.08} & \textbf{93.80$\pm$0.08} & 95.28$\pm$0.05 & 93.58$\pm$0.05 & \textbf{95.33$\pm$0.03} & \textbf{93.62$\pm$0.03} \\
ro & 91.97$\pm$0.13 & 85.38$\pm$0.03 & \textbf{92.06$\pm$0.08} & \textbf{85.58$\pm$0.12} & \textbf{91.94$\pm$0.07} & \textbf{85.61$\pm$0.13} & 91.80$\pm$0.11 & 85.34$\pm$0.21 \\
ru & 93.81$\pm$0.05 & 91.85$\pm$0.06 & \textbf{94.11$\pm$0.07} & \textbf{92.29$\pm$0.10} & 94.40$\pm$0.03 & 92.68$\pm$0.04 & \textbf{94.69$\pm$0.04} & \textbf{93.07$\pm$0.03} \\
\hline
\end{tabular}
}
\caption{UAS and LAS on both the development and test datasets of 12 treebanks from UD Treebanks, together with \textsc{BiAF} for comparison.}
\label{tab:udt}
\end{table*}

\subsection{Experiments on Other Treebanks}\label{subsec:experiment:other}
\subsubsection{CoNLL Treebanks}
Table~\ref{tab:conll} summarizes the parsing results of our model on the test sets of 14 treebanks from the CoNLL shared task, along with the state-of-the-art
baselines. 
Along with \textsc{BiAF}, we also list the performance of the bi-directional attention based Parser (\textsf{Bi-Att}) \citep{cheng-EtAl:2016:EMNLP2016} and the neural MST parser (\textsf{NeuroMST}) \citep{ma-hovy:2017:I17-1} for comparison. 
Our parser achieves state-of-the-art performance on both UAS and LAS on 
eight languages --- Arabic, Czech, English, German, Portuguese, Slovene, Spanish, and Swedish. 
On Bulgarian and Dutch, our parser obtains the best UAS.
On other languages, the performance of our parser is competitive with \textsc{BiAF}, and significantly better than others.
The only exception is Japanese, on which \textsf{NeuroMST} obtains the best scores.

\subsubsection{UD Treebanks} 
For UD Treebanks, we select 12 languages --- Bulgarian, Catalan, Czech, Dutch, English, French, German, Italian, Norwegian, Romanian, Russian and Spanish. 
For all the languages, we adopt the standard training/dev/test splits, and use the universal POS tags~\citep{PETROV12.274.L12-1115} provided in each treebank. 
The statistics of these corpora are provided in Appendix B.

Table~\ref{tab:udt} summarizes the results of the \textsc{StackPtr} parser, along with \textsc{BiAF} for comparison, on both the development and test datasets for each language.
First, both \textsc{BiAF} and \textsc{StackPtr} parsers achieve relatively high parsing accuracies on all the 12 languages --- all with UAS are higher than 90\%. 
On nine languages --- Catalan, Czech, Dutch, English, French, German, Norwegian, Russian and Spanish --- \textsc{StackPtr} outperforms \textsc{BiAF} for both UAS and LAS. 
On Bulgarian, \textsc{StackPtr} achieves slightly better UAS while LAS is slightly worse than \textsc{BiAF}.
On Italian and Romanian, \textsc{BiAF} obtains marginally better parsing performance than \textsc{StackPtr}.

\section{Conclusion}
In this paper, we proposed \textsc{StackPtr}, a transition-based neural network architecture, for dependency parsing.
Combining pointer networks with an internal stack to track the status of the top-down, depth-first search in the decoding procedure, the \textsc{StackPtr} parser is able to capture information from the whole sentence and all the previously derived subtrees, removing the left-to-right restriction in classical transition-based parsers, while maintaining linear parsing steps, w.r.t the length of the sentences.
Experimental results on 29 treebanks show the effectiveness of our parser across 20 languages, by achieving state-of-the-art performance on 21 corpora.

There are several potential directions for future work.
First, we intend to consider how to conduct experiments to improve the analysis of parsing errors qualitatively and quantitatively. 
Another interesting direction is to further improve our model by exploring reinforcement learning approaches to learn an optimal order for the children of head words, instead of using a predefined fixed order.

\section*{Acknowledgements}
The authors thank Chunting Zhou, Di Wang and Zhengzhong Liu for their helpful discussions.
This research was supported in part by DARPA
grant FA8750-18-2-0018 funded under the AIDA
program. Any opinions, findings, and conclusions
or recommendations expressed in this material are
those of the authors and do not necessarily reflect
the views of DARPA.

\bibliography{acl2018}
\bibliographystyle{acl_natbib}

\newpage
\onecolumn
{\centering \large {\bf Appendix: Stack-Pointer Network for Dependency Parsing}} \\
\section*{Appendix A: Hyper-Parameters}
Table~\ref{tab:hyper-params} summarizes the chosen hyper-parameters used for all the experiments in this paper.
Some parameters are chosen directly or similarly from those reported in \citet{dozat2017:ICLR}.
We use the same hyper-parameters across the models on different treebanks and languages, due to time constraints.
\begin{table}[h]
\centering
\begin{tabular}[t]{l|l|r}
\hline
\textbf{Layer} & \textbf{Hyper-parameter} & \textbf{Value} \\
\hline
\multirow{2}{*}{CNN} & window size & 3 \\
 & number of filters & 50 \\
\hline
\multirow{4}{*}{LSTM} & encoder layers & 3 \\
 & encoder size & 512 \\
\cline{2-3}
 & decoder layers & 1 \\
 & decoder size & 512 \\
\hline
\multirow{2}{*}{MLP} & arc MLP size & 512 \\
 & label MLP size & 128 \\
\hline
\multirow{3}{*}{Dropout} & embeddings & 0.33 \\
 & LSTM hidden states & 0.33 \\
 & LSTM layers & 0.33 \\
\hline
\multirow{5}{*}{Learning} & optimizer & Adam \\
 & initial learning rate & 0.001 \\
 & $(\beta_1, \beta_2)$ & (0.9, 0.9) \\
 & decay rate & 0.75 \\
 & gradient clipping & 5.0 \\
\hline
\end{tabular}
\caption{Hyper-parameters for all experiments.}
\label{tab:hyper-params}
\end{table}

\newpage
\section*{Appendix B: UD Treebanks}
Table~\ref{tab:udt:stats} shows the corpora statistics of the treebanks for 12 languages. For evaluation, we report results excluding punctuation, which is any tokens with POS tags ``PUNCT'' or ``SYM''.
\begin{table}[h]
\centering
\begin{tabular}[t]{l|l|cc:c}
\hline
 & Corpora & & \emph{\#Sent} & \emph{\#Token (w.o punct)} \\
\hline
\multirow{3}{*}{Bulgarian} & & Training & 8,907 & 124,336 (106,813) \\
 & BTB & Dev & 1,115 & 16,089 (13,822) \\
 & & Test & 1,116 & 15,724 (13,456) \\
\hline
\multirow{3}{*}{Catalan} & & Training & 13,123 & 417,587 (371,981) \\
& AnCora & Dev & 1,709 & 56,482 (50,452) \\
& & Test & 1,846 & 57,738 (51,324) \\
\hline
\multirow{3}{*}{Czech} & PDT, CAC & Training & 102,993 & 1,806,230 (1,542,805) \\
& CLTT & Dev & 11,311 & 191,679 (163,387) \\
& FicTree & Test & 12,203 & 205,597 (174,771) \\
\hline
\multirow{3}{*}{Dutch} & Alpino & Training & 18,310 & 267,289 (234,104) \\
& LassySmall & Dev & 1,518 & 22,091 (19,042)	\\
& & Test & 1,396 & 21,126 (18,310) \\
\hline
\multirow{3}{*}{English} & & Training & 12,543 & 204,585 (180,308) \\
& EWT & Dev & 2,002 & 25,148 (21,998) \\
& & Test & 2,077 & 25,096 (21,898)\\
\hline
\multirow{3}{*}{French} & & Training & 14,554 & 356,638 (316,780) \\
& GSD & Dev & 1,478 & 35,768 (31,896) \\
& & Test & 416 & 10,020 (8,795) \\
\hline
\multirow{3}{*}{German} & & Training & 13,841 & 263,536 (229,204) \\
& GSD & Dev & 799 & 12,348 (10,727) \\
& & Test & 977 & 16,268 (13,929) \\
\hline
\multirow{3}{*}{Italian} & & Training & 12,838 & 270,703 (239,836) \\
& ISDT & Dev & 564 & 11,908 (10,490) \\
& & Test & 482 & 10,417 (9,237) \\
\hline
\multirow{3}{*}{Norwegian} & Bokmaal & Training & 29,870 & 48,9217 (43,2597) \\
& Nynorsk & Dev & 4,300 & 67,619 (59,784) \\
& & Test & 3,450 & 54,739 (48,588)	\\
\hline
\multirow{3}{*}{Romanian} & & Training & 8,043 & 185,113 (161,429) \\
& RRT & Dev & 752 & 17,074 (14,851) \\
& & Test & 729 & 16,324 (14,241) \\
\hline
\multirow{3}{*}{Russian} & & Training & 48,814 & 870,034 (711,184) \\
& SynTagRus & Dev & 6,584 & 118,426 (95,676) \\
& & Test & 6,491 & 117,276 (95,745) \\
\hline
\multirow{3}{*}{Spanish} & GSD & Training & 28,492 & 827,053 (730,062) \\
& AnCora & Dev & 4,300 & 89,487 (78,951) \\
& & Test & 2,174 & 64,617 (56,973) \\
\hline
\end{tabular}
\caption{Corpora statistics of UD Treebanks for 12 languages.
\emph{\#Sent} and \emph{\#Token} refer to the number of sentences and the number of words (w./w.o punctuations) in each data set, respectively.}
\label{tab:udt:stats}
\end{table}

\newpage
\section*{Appendix C: Main Results}
Table~\ref{tab:main:results} illustrates the details of the experimental results.
For each \textsc{StackPrt} parsing model, we ran experiments with decoding beam size equals to 1, 5, and 10.
For each experiment, we report the mean values with corresponding standard deviations over 5 runs.
\begin{table*}[h]
\centering
\setlength{\tabcolsep}{3pt}
{\small
\begin{tabular}[t]{l|c|cccc|cccc}
\hline
& & \multicolumn{8}{c}{\textbf{English}} \\
\cline{3-10}
& & \multicolumn{4}{c|}{\textbf{Dev}} & \multicolumn{4}{c}{\textbf{Test}} \\
\cline{3-10}
\textbf{Model} & beam & UAS & LAS & UCM & LCM & UAS & LAS & UCM & LCM \\
\hline
BiAF & -- & 95.73$\pm$0.04 & \textbf{93.97$\pm$0.06} & 60.58$\pm$0.77 & 47.47$\pm$0.63 & 95.84$\pm$0.06 & \textbf{94.21$\pm$0.04} & 59.49$\pm$0.23 & 49.07$\pm$0.34 \\
\hline
\hline
\multirow{3}{*}{\textsf{Basic}} & 1 & 95.71$\pm$0.02 & 93.88$\pm$0.03 & 62.33$\pm$0.33 & 47.75$\pm$0.32 & 95.71$\pm$0.06 & 94.07$\pm$0.06 & 60.91$\pm$0.35 & 49.54$\pm$0.48 \\
& 5 & 95.71$\pm$0.04 & 93.88$\pm$0.05 & 62.40$\pm$0.45 & 47.80$\pm$0.44 & 95.76$\pm$0.11 & 94.12$\pm$0.11 & 61.09$\pm$0.43 & 49.67$\pm$0.41 \\
& 10 & 95.72$\pm$0.03 & 93.89$\pm$0.04 & \textbf{62.40$\pm$0.45} & \textbf{47.80$\pm$0.44} & 95.77$\pm$0.11 & 94.12$\pm$0.11 & 61.09$\pm$0.43 & 49.67$\pm$0.41 \\
\hline
\multirow{3}{*}{\textsf{+gpar}} & 1 & 95.68$\pm$0.04 & 93.82$\pm$0.02 & 61.82$\pm$0.36 & 47.32$\pm$0.14 & 95.73$\pm$0.04 & 94.07$\pm$0.05 & 60.99$\pm$0.46 & 49.83$\pm$0.59 \\
& 5 & 95.67$\pm$0.01 & 93.83$\pm$0.02 & 61.93$\pm$0.32 & 47.44$\pm$0.20 & 95.76$\pm$0.06 & 94.11$\pm$0.06 & 61.23$\pm$0.47 & 50.07$\pm$0.59 \\
& 10 & 95.69$\pm$0.02 & 93.83$\pm$0.02 & 61.95$\pm$0.32 & 47.44$\pm$0.20 & 95.78$\pm$0.05 & 94.12$\pm$0.06 & 61.24$\pm$0.46 & \textbf{50.07$\pm$0.59} \\
\hline
\multirow{3}{*}{\textsf{+sib}} & 1 & 95.75$\pm$0.03 & 93.93$\pm$0.04 & 61.93$\pm$0.49 & 47.66$\pm$0.48 & 95.77$\pm$0.15 & 94.11$\pm$0.06 & 61.32$\pm$0.37 & 49.75$\pm$0.29 \\
& 5 & 95.74$\pm$0.02 & 93.93$\pm$0.05 & 62.16$\pm$0.22 & 47.68$\pm$0.54 & 95.84$\pm$0.09 & 94.17$\pm$0.09 & 61.52$\pm$0.57 & 49.91$\pm$0.76 \\
& 10 & 95.75$\pm$0.02 & 93.94$\pm$0.06 & 62.17$\pm$0.20 & 47.68$\pm$0.54 & 95.85$\pm$0.10 & 94.18$\pm$0.09 & \textbf{61.52$\pm$0.57} & 49.91$\pm$0.76 \\
\hline
\multirow{3}{*}{\textsf{Full}} & 1 & 95.63$\pm$0.08 & 93.78$\pm$0.08 & 61.56$\pm$0.63 & 47.12$\pm$0.36 & 95.79$\pm$0.06 & 94.11$\pm$0.06 & 61.02$\pm$0.31 & 49.45$\pm$0.23 \\
& 5 & 95.75$\pm$0.06 & 93.90$\pm$0.08 & 62.06$\pm$0.42 & 47.43$\pm$0.36 & 95.87$\pm$0.04 & 94.20$\pm$0.03 & 61.43$\pm$0.49 & 49.68$\pm$0.47 \\
& 10 & \textbf{95.75$\pm$0.06} & 93.90$\pm$0.08 & 62.08$\pm$0.39 & 47.43$\pm$0.36 & \textbf{95.87$\pm$0.04} & 94.19$\pm$0.04 & 61.43$\pm$0.49 & 49.68$\pm$0.47 \\
\hline
& & \multicolumn{8}{c}{\textbf{Chinese}} \\
\cline{3-10}
& & \multicolumn{4}{c|}{\textbf{Dev}} & \multicolumn{4}{c}{\textbf{Test}} \\
\cline{3-10}
\textbf{Model} & beam & UAS & LAS & UCM & LCM & UAS & LAS & UCM & LCM \\
\hline
BiAF & -- & 90.20$\pm$0.17 & 88.94$\pm$0.13 & 43.41$\pm$0.83 & 38.42$\pm$0.79 & 90.43$\pm$0.08 & 89.14$\pm$0.09 & 42.92$\pm$0.29 & 38.68$\pm$0.25 \\
\hline
\hline
\multirow{3}{*}{\textsf{Basic}} & 1 & 89.76$\pm$0.32 & 88.44$\pm$0.28 & 45.18$\pm$0.80 & 40.13$\pm$0.63 & 90.04$\pm$0.32 & 88.74$\pm$0.40 & 45.00$\pm$0.47 & 40.12$\pm$0.42 \\
& 5 & 89.97$\pm$0.13 & 88.67$\pm$0.14 & 45.33$\pm$0.58 & 40.25$\pm$0.65 & 90.46$\pm$0.15 & 89.17$\pm$0.18 & 45.41$\pm$0.48 & 40.53$\pm$0.48 \\
& 10 & 89.97$\pm$0.14 & 88.68$\pm$0.14 & 45.33$\pm$0.58 & 40.25$\pm$0.65 & 90.48$\pm$0.11 & 89.19$\pm$0.15 & 45.44$\pm$0.44 & 40.56$\pm$0.43 \\
\hline
\multirow{3}{*}{\textsf{+gpar}} & 1 & 90.05$\pm$0.14 & 88.71$\pm$0.16 & 45.63$\pm$0.52 & 40.45$\pm$0.61 & 90.28$\pm$0.10 & 88.96$\pm$0.10 & 45.26$\pm$0.59 & 40.38$\pm$0.43 \\
& 5 & 90.17$\pm$0.14 & 88.85$\pm$0.13 & 46.03$\pm$0.53 & 40.69$\pm$0.55 & 90.45$\pm$0.15 & 89.14$\pm$0.14 & 45.71$\pm$0.46 & 40.80$\pm$0.26 \\
& 10 & 90.18$\pm$0.16 & 88.87$\pm$0.14 & \textbf{46.05$\pm$0.58} & 40.69$\pm$0.55 & 90.46$\pm$0.16 & 89.16$\pm$0.15 & 45.71$\pm$0.46 & 40.80$\pm$0.26 \\
\hline
\multirow{3}{*}{\textsf{+sib}}& 1 & 89.91$\pm$0.07 & 88.59$\pm$0.10 & 45.50$\pm$0.50 & 40.40$\pm$0.48 & 90.25$\pm$0.10 & 88.94$\pm$0.12 & 45.42$\pm$0.52 & 40.54$\pm$0.69 \\
& 5 & 89.99$\pm$0.05 & 88.70$\pm$0.09 & 45.55$\pm$0.36 & 40.37$\pm$0.14 & 90.41$\pm$0.07 & 89.12$\pm$0.07 & 45.76$\pm$0.46 & 40.69$\pm$0.52 \\
& 10 & 90.00$\pm$0.04 & 88.72$\pm$0.09 & 45.58$\pm$0.32 & 40.37$\pm$0.14 & 90.43$\pm$0.09 & 89.15$\pm$0.10 & 45.75$\pm$0.44 & 40.68$\pm$0.50 \\
\hline
\multirow{3}{*}{\textsf{Full}} & 1 & 90.21$\pm$0.15 & 88.85$\pm$0.15 & 45.83$\pm$0.52 & 40.54$\pm$0.60 & 90.36$\pm$0.16 & 89.05$\pm$0.15 & 45.60$\pm$0.33 & 40.73$\pm$0.23 \\
& 5 & 90.23$\pm$0.13 & 88.89$\pm$0.14 & 46.00$\pm$0.54 & \textbf{40.75$\pm$0.64} & 90.58$\pm$0.12 & 89.27$\pm$0.11 & \textbf{46.20$\pm$0.26} & \textbf{41.25$\pm$0.22} \\
& 10 & \textbf{90.29$\pm$0.13} & \textbf{88.95$\pm$0.13} & 46.03$\pm$0.54 & \textbf{40.75$\pm$0.64} & \textbf{90.59$\pm$0.12} & \textbf{89.29$\pm$0.11} & \textbf{46.20$\pm$0.26} & \textbf{41.25$\pm$0.22} \\
\hline
& & \multicolumn{8}{c}{\textbf{German}} \\
\cline{3-10}
& & \multicolumn{4}{c|}{\textbf{Dev}} & \multicolumn{4}{c}{\textbf{Test}} \\
\cline{3-10}
\textbf{Model} & beam & UAS & LAS & UCM & LCM & UAS & LAS & UCM & LCM \\
\hline
BiAF & -- & \textbf{93.60$\pm$0.13} & \textbf{91.96$\pm$0.13} & 58.79$\pm$0.25 & 49.59$\pm$0.19 & \textbf{93.85$\pm$0.07} & \textbf{92.32$\pm$0.06} & 60.60$\pm$0.38 & 52.46$\pm$0.24 \\
\hline
\hline
\multirow{3}{*}{\textsf{Basic}} & 1 & 93.35$\pm$0.14 & 91.58$\pm$0.17 & 59.64$\pm$0.78 & 49.75$\pm$0.64 & 93.39$\pm$0.09 & 91.85$\pm$0.09 & 61.08$\pm$0.31 & 52.21$\pm$0.53 \\
& 5 & 93.49$\pm$0.14 & 91.72$\pm$0.16 & 59.99$\pm$0.69 & 49.82$\pm$0.54 & 93.61$\pm$0.09 & 92.07$\pm$0.08 & 61.38$\pm$0.30 & 52.51$\pm$0.43 \\
& 10 & 93.48$\pm$0.14 & 91.71$\pm$0.17 & \textbf{60.02$\pm$0.69} & 49.84$\pm$0.54 & 93.59$\pm$0.09 & 92.06$\pm$0.08 & 61.38$\pm$0.30 & 52.51$\pm$0.43 \\
\hline
\multirow{3}{*}{\textsf{+gpar}} & 1 & 93.39$\pm$0.07 & 91.66$\pm$0.13 & 59.59$\pm$0.54 & 49.81$\pm$0.42 & 93.44$\pm$0.07 & 91.91$\pm$0.11 & 61.73$\pm$0.47 & 52.84$\pm$0.48 \\
& 5 & 93.47$\pm$0.09 & 91.75$\pm$0.10 & 59.81$\pm$0.55 & 50.05$\pm$0.39 & 93.68$\pm$0.04 & 92.16$\pm$0.04 & 62.09$\pm$0.44 & 53.13$\pm$0.42 \\
& 10 & 93.48$\pm$0.08 & 91.76$\pm$0.09 & 59.89$\pm$0.59 & 50.09$\pm$0.40 & 93.68$\pm$0.05 & 92.16$\pm$0.03 & 62.10$\pm$0.42 & \textbf{53.14$\pm$0.4} \\
\hline
\multirow{3}{*}{\textsf{+sib}} & 1 & 93.43$\pm$0.07 & 91.73$\pm$0.08 & 59.68$\pm$0.25 & 49.93$\pm$0.30 & 93.55$\pm$0.07 & 92.00$\pm$0.08 & 61.90$\pm$0.50 & 52.79$\pm$0.22 \\
& 5 & 93.53$\pm$0.05 & 91.83$\pm$0.07 & 59.95$\pm$0.23 & 50.14$\pm$0.39 & 93.75$\pm$0.09 & 92.20$\pm$0.08 & 62.21$\pm$0.38 & 53.03$\pm$0.18 \\
& 10 & 93.55$\pm$0.06 & 91.84$\pm$0.07 & 59.96$\pm$0.24 & \textbf{50.15$\pm$0.40} & 93.76$\pm$0.09 & 92.21$\pm$0.08 & \textbf{62.21$\pm$0.38} & 53.03$\pm$0.18 \\
\hline
\multirow{3}{*}{\textsf{Full}} & 1 & 93.33$\pm$0.13 & 91.60$\pm$0.16 & 59.78$\pm$0.32 & 49.78$\pm$0.29 & 93.50$\pm$0.04 & 91.91$\pm$0.11 & 61.80$\pm$0.28 & 52.95$\pm$0.37 \\
& 5 & 93.42$\pm$0.11 & 91.69$\pm$0.12 & 59.90$\pm$0.27 & 49.94$\pm$0.35 & 93.64$\pm$0.03 & 92.10$\pm$0.06 & 61.89$\pm$0.21 & 53.06$\pm$0.36 \\
& 10 & 93.40$\pm$0.11 & 91.67$\pm$0.12 & 59.90$\pm$0.27 & 49.94$\pm$0.35 & 93.64$\pm$0.03 & 92.11$\pm$0.05 & 61.89$\pm$0.21 & 53.06$\pm$0.36 \\
\hline
\end{tabular}
}
\caption{Parsing performance of different variations of our model on both the development and test sets for three languages, together with the \textsc{BiAF} parser as the baseline. Best results are highlighted with bold print.}
\label{tab:main:results}
\end{table*}

\end{document}